\pdfoutput=1
\documentclass[runningheads]{llncs}
\usepackage{cite}
\usepackage{amsmath,amssymb,amsfonts}
\usepackage{algorithmic}
\usepackage{graphicx}
\usepackage{textcomp}
\usepackage{xcolor}
\usepackage{subfig}
\usepackage{listings}
\usepackage{pythonhighlight}

\usepackage[ruled]{algorithm2e} %for psuedo code

\newcommand{\dataNum}{N} % Num of data
\newcommand{\timeInputStep}{s} % time step
\newcommand{\timeOutputStep}{t} % time step
\newcommand{\dataStep}{n} % data step
\newcommand{\seqInputLen}{S} % input sequence length
\newcommand{\seqOutputLen}{T} % output sequence length
\newcommand{\inputSeq}{x_{1:\seqInputLen}} % input sequence
\newcommand{\inputStep}{x} % input step
\newcommand{\outputSeqVec}{y_{1:\seqOutputLen}} % output sequence
\newcommand{\outputStep}{y} % output step
\newcommand{\hiddenEncStep}{\mathbf{h}^\mathrm{enc}}
\newcommand{\hiddenDecStep}{\mathbf{h}^\mathrm{dec}}
\newcommand{\contextStep}{\mathbf{c}}
\newcommand{\repEncStep}{\mathbf{r}^\mathrm{enc}}
\newcommand{\repDecStep}{\mathbf{r}^\mathrm{dec}}
\newcommand{\windowEncStep}{W^\mathrm{enc}}
\newcommand{\windowDecStep}{W^\mathrm{dec}}
\newcommand{\alignmentStep}{a}
\newcommand{\param}{\mathbf{\theta}}
\newcommand{\calD}{\mathcal{D}}
\newcommand{\calL}{\mathcal{L}}
\newcommand{\score}{\mbox{score}}
\newcommand{\loss}{\calL(\param)}
\newcommand{\inputSeqVec}{\mathbf{x}_{1:\seqInputLen}} % input sequence
\newcommand{\outputSeqVecD}{\mathbf{y}_{1:\seqOutputLen}} % output sequence
\newcommand{\outputStepVec}{\mathbf{y}} % output step
\newcommand{\inputStepVec}{\mathbf{x}} % input step

\newcommand{\outputFunc}{g} % output function 

\newcommand{\windowEncWidth}{w^\mathrm{enc}}
\newcommand{\windowDecWidth}{w^\mathrm{dec}}

\newcommand{\dimInput}{d_\mathbf{x}}
\newcommand{\dimOutput}{d_\mathbf{y}}
\newcommand{\dimRep}{d_\mathbf{r}}

\newcommand{\timeInputWinStep}{{\timeInputStep'}}
\newcommand{\timeOutputWinStep}{{\timeOutputStep'}}
\newcommand{\seqInputWinLen}{S'} % input sequence length
\newcommand{\seqOutputWinLen}{T'} % output sequence length

\newcommand{\repFunc}{\mathrm{R}}
\newcommand{\invRepFunc}{\mathrm{R}^{-1}}
\newcommand{\dimZ}{d_\mathbf{z}}
\def\BibTeX{{\rm B\kern-.05em{\sc i\kern-.025em b}\kern-.08emT\kern-.1667em\lower.7ex\hbox{E}\kern-.125emX}}

\makeatletter
\def\blfootnote{\xdef\@thefnmark{}\@footnotetext}
\makeatother

\begin{document}
\lstset{language=Python}
\lstset{frame=lines}
\lstset{caption={Insert code directly in your document}}
\lstset{label={lst:code_direct}}
\lstset{basicstyle=\footnotesize}

\title{Translation Between Waves, {\it wave2wave}}

% If the paper title is too long for the running head, you can set
% an abbreviated paper title here
%
\author{Tsuyoshi Okita$^\dagger$\inst{1,2} \and
  Hirotaka Hachiya$^\ddagger$\inst{3,2} \and
  Sozo Inoue\inst{1,2} \and
  Naonori Ueda\inst{2}
}

\authorrunning{T. Okita et al.}
%\authorrunning{F. Author et al.}
% First names are abbreviated in the running head.
% If there are more than two authors, 'et al.' is used.
%
\institute{Kyushu Institute of Technology \and Riken AIP \and Wakayama University\\
  \email{\{tsuyoshi.okita, hirotaka.hachiya, sozo.inoue, naonori.ueda\}@riken.jp}}
%\institute{Princeton University, Princeton NJ 08544, USA \and
%Springer Heidelberg, Tiergartenstr. 17, 69121 Heidelberg, Germany
%\email{lncs@springer.com}\\
%\url{http://www.springer.com/gp/computer-science/lncs} \and
%ABC Institute, Rupert-Karls-University Heidelberg, Heidelberg, Germany\\
%\email{\{abc,lncs\}@uni-heidelberg.de}}
%
\maketitle              % typeset the header of the contribution

\vspace{-0.5cm}
\begin{abstract}
  The understanding of sensor data has been greatly improved by
  advanced deep learning methods with big data. However, available
  sensor data in the real world are still limited, which is called the
  opportunistic sensor problem.  This paper proposes a new variant of
  neural machine translation {\it seq2seq} to deal with continuous
  signal waves by introducing the {\it window-based (inverse-) representation} to adaptively represent partial shapes of waves
  and the {\it iterative back-translation model} for high-dimensional
  data.  Experimental results are shown for two real-life data:
  earthquake and activity translation.  The performance improvements
  of one-dimensional data was about 46 \% in test loss and that of
  high-dimensional data was about 1625 \% in perplexity with regard to the original seq2seq. 
  %\keywords{sequence to sequence models \and deep learning \and spatio-temporal model \and earthquake translation \and activity translation.}
\end{abstract}

\vspace{-0.8cm}
\section{Introduction}\blfootnote{$\dagger$ and $\ddagger$ contributed equally.}

The problem of shortage of training data but can be supplied by other
sensor data is called an {\it opportunistic sensor problem}
\cite{Roggen13}.  For example in human activity logs, the video data
can be missing in bathrooms by ethical reasons but can be supplied by
environmental sensors which have less ethical problems. For this
purpose we propose to extend the sequence-to-sequence (seq2seq) model
\cite{Luong15} to translate signal wave $x$ ({\it continuous
  time-series signals}) into other signal wave $y$.  The
straight-forward extension does not apply by two reasons: (1) the
lengths of $x$ and $y$ are radically different, and (2) both $x$ and
$y$ are high dimensions.

First, while most of the conventional seq2seq models handle the input
and output signals whose lengths are in the same order, we need to
handle the output signals whose length are sometimes considerably
different than the input signals. For example, the sampling rate of
ground motion sensor is $100\mathbf{Hz}$ and the duration of an
earthquake is about $10\mathbf{sec}$. That is, the length of the
output signal wave is $10000$ times longer in this case.  Therefore,
the segmentation along temporal axis and discarding uninformative
signal waves are required.  Second, signal waves could be high
dimensions; motion capture data is in $129$-dimensionals and
acceleormeter data is in $18$-dimensionals.  While most of the
conventional seq2seq does not require the high dimensional settings,
meaning that it is not usual to translate multiple languages
simultaneously, we need to translate signal waves in high dimensions
into other signal waves in high dimensions simultaneously.

To overcome these two problems we propose 1) the window-based
representation function and 2) the {\it wave2wave} iterative
back-translation model in this paper.  Our contributions are the
following:
\begin{itemize}
\item We propose a sliding window-based seq2seq model {\it wave2wave} (Section \ref{WindowBasedRepresentation}),
\item We propose the {\it wave2wave} iterative back-translation model (Section \ref{Wave2waveIterative}) which is the key to outperform for high-dimensional data.
\end{itemize}

\vspace{-0.8cm}
\section{seq2seq}

\paragraph{Architecture with context vector}
Let $\inputSeq$ $=(\inputStep_1,\inputStep_2, \ldots,\inputStep_\seqInputLen)$ denotes a source sentence consisting of time-series $\seqInputLen$ words, 
and $\outputSeqVec = (\outputStep_1,\dots, \outputStep_\seqOutputLen)$ denotes a target sentence corresponding to $\inputSeq$.
With the assumption of a Markov property, the conditional probability $p(\outputSeqVec|\inputSeq)$, translation from a source sentence to a target sentence, is decomposed into a time-step translation $p(\outputStep|\inputStep)$ as in
$\log p(\outputSeqVec|\inputSeq) = \sum_{\timeOutputStep=1}^\seqOutputLen \log p(\outputStep_\timeOutputStep | \outputStep_{<\timeOutputStep},\contextStep_\timeOutputStep)$
\label{conditionalProb}
where $\outputStep_{<\timeInputStep}=(\outputStep_1,\outputStep_2,\ldots,\outputStep_{\timeInputStep-1})$ and $\contextStep_\timeInputStep$ is a context vector representing the information of source sentence $\inputSeq$ to generate an output word $\outputStep_\timeOutputStep$. 

To realize such time-step translation, the seq2seq architecture consists of (a) a RNN (Reccurent Neural Network) encoder
and (b) a RNN decoder. The RNN encoder computes the current hidden state $\hiddenEncStep_\timeInputStep$ given the previous hidden state $\hiddenEncStep_{\timeInputStep-1}$ and the current input $\inputStep_\timeInputStep$,
as in $\hiddenEncStep_\timeInputStep = \mbox{RNN}_{\mbox{enc}}(\inputStep_\timeInputStep, \hiddenEncStep_{\timeInputStep-1})$
  \label{RNNenc}
where $\mbox{RNN}_{\mbox{enc}}$ denotes a multi-layered RNN unit.
The RNN decoder computes a current hidden state $\hiddenDecStep_\timeOutputStep$ given the
previous hidden state and then compute an output $\outputStep_\timeOutputStep$ by
$\hiddenDecStep_{\timeOutputStep} = \mbox{RNN}_{\mbox{dec}}(\hiddenDecStep_{\timeOutputStep-1})$ and 
$p_\param(\outputStep_\timeOutputStep | \outputStep_{<\timeOutputStep},\contextStep_\timeOutputStep) = \mbox{softmax} \Big(\outputFunc_\param(\hiddenDecStep_{\timeOutputStep},\contextStep_{\timeOutputStep})\Big)$
\label{RNNdec}
where $\mbox{RNN}_{\mbox{dec}}$ denotes a conditional RNN unit, $\outputFunc_\param(\cdot)$ is the output function to convert $\hiddenDecStep_\timeOutputStep$ and $\contextStep_\timeOutputStep$ to the logit of  $\outputStep_\timeOutputStep$, and $\param$ denotes parameters in RNN units.

With training data $\calD=\{\outputSeqVec^\dataStep,\inputSeq^\dataStep\}_{\dataStep=1}^\dataNum$, the parameters $\param$ are optimized so as to minimize the loss function %$\loss$ of log-likelihood, as in
of log-likelihood
%\begin{equation}
$\calL(\param) (= -\frac{1}{\dataNum}\sum_{\dataStep=1}^{\dataNum} \sum_{\timeOutputStep=1}^{\seqOutputLen} \log p_\param$ $(\outputStep^\dataStep_\timeOutputStep|\outputStep^\dataStep_{<\timeOutputStep},\contextStep_\timeOutputStep)$)
\label{lossSeq2seq}
or squared error
$\calL(\param) (= \frac{1}{\dataNum}\sum_{\dataStep=1}^{\dataNum}\sum_{\timeOutputStep=1}^{\seqOutputLen}
\Big(\outputStep^{\dataStep}_\timeOutputStep -
\outputFunc_\param (
{{\hiddenDecStep}^{\dataStep}}_{\timeOutputStep},
\contextStep^{\dataStep}_{\timeOutputStep}
)
\Big)^2$).

\paragraph{Global Attention}
To obtain the context vector $\contextStep_\timeInputStep$, we use global attention mechanism~\cite{Luong15}.
The global attention considers an 
attention mapping in a global manner,
between encoder hidden states
$\hiddenEncStep_\timeInputStep$ and a decoder hidden step $\hiddenDecStep_\timeOutputStep$ by
  $\alignmentStep_\timeOutputStep(\timeInputStep) = \mbox{align}  (\hiddenDecStep_\timeOutputStep, \hiddenEncStep_\timeInputStep) \label{energy}
  = \frac{\exp(\score(\hiddenDecStep_\timeOutputStep, \hiddenEncStep_\timeInputStep)}{\sum_\timeInputStep^\seqOutputLen \exp (\score(\hiddenDecStep_\timeOutputStep, \hiddenEncStep_\timeInputStep))}$,
where the score is computed by weighted inner product by
   $\score(\hiddenDecStep_\timeOutputStep,\hiddenEncStep_\timeInputStep) =
   {{\hiddenDecStep}_{\timeOutputStep}}^\top W_\alignmentStep \hiddenEncStep_\timeInputStep$, 
\label{scoreAttention}
where the weight parameter $W_\alignmentStep$ is obtained so as to minimize the loss function $\loss$.
Then, the context vector $\contextStep_\timeOutputStep$ is obtained as a weighted average of encoder hidden states
by
    $\contextStep_\timeOutputStep = \sum_{\timeInputStep=1}^\seqInputLen
    \alignmentStep_\timeOutputStep(\timeInputStep) \hiddenEncStep_\timeInputStep$

\vspace{-0.3cm}
\section{Proposed method: wave2wave}
\begin{figure*}[t]
\begin{minipage}{1\textwidth}
 \begin{center}
 \includegraphics[width=0.9\textwidth]{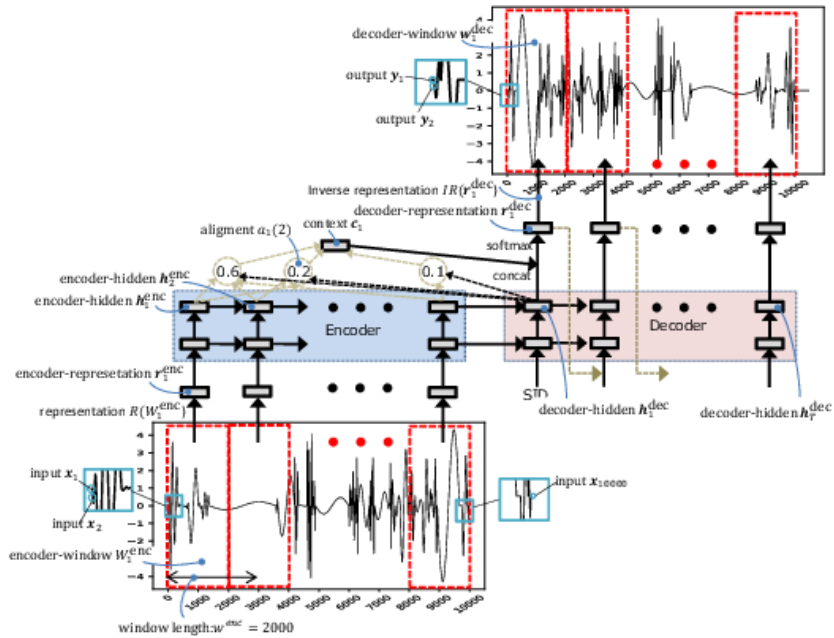}   
 \end{center}
\end{minipage}
 \caption{Overall architecture of our method, wave2wave, consisting RNN encoder and decoder with context vector and sliding window representation. Input and output time-series data are toy examples where the input is generated by combining sine waves with random magnitudes and pepriods. The output is the version of the input flipped horizontally.}
 \label{fig:wave2wave-architecture}
\end{figure*}
\vspace{-0.2cm}

The problems of global attention model are that (1) the lengths of
input and output are radically different, and that (2) both input and
output sequences are high dimensionals.
For example in activity translation, there are $48$ motion sensors and $3$ accelerometer sensors.
Their frequency rates are as high as $50\mathrm{Hz}$ and $30\mathrm{Hz}$ respectively.
Therefore, the number of steps $\seqInputLen$, $\seqOutputLen$ in both encoder and decoders 
are prohibitively large so that the capturing information of source sentence $\inputSeq$ is precluded in the context vector $\contextStep$.

\vspace{-0.3cm}
\subsection{Window-based representation}\label{WindowBasedRepresentation}
Let us consider the case that source and target sentences are multi-dimensional continuous time-series, signal waves, as shown in Fig.~\ref{fig:wave2wave-architecture}~\footnote{We note that signal waves in Fig.~\ref{fig:wave2wave-architecture} are depicted as one-dimensional waves for clear visualization.}.
That is,  each signal at time-step $\inputSeq$ is expressed as 
$\dimInput$-dimensional vector $\inputStepVec_\timeInputStep$---there are $\dimInput$ sensors in the source side.
Then a source signal wave $\inputSeqVec$ consists of  
$\seqInputLen$-step $\dimInput$-dimensional signal vectors, 
i.e., $\inputSeqVec=(\inputStepVec_1,\inputStepVec_2, \ldots,\inputStepVec_\seqInputLen)$.

To capture an important shape informaion from complex signal waves (see Fig.~\ref{fig:wave2wave-architecture}), we introduce trainable window-based representation function $\repFunc(\cdot)$ as
\begin{equation}
   \repEncStep_\timeInputWinStep = \repFunc(\windowEncStep_\timeInputWinStep)
   \label{eq:encoder-representation}
\end{equation}
where $\windowEncStep_\timeInputWinStep$ is a $\timeInputWinStep$-th window with fixed window-width $\windowEncWidth$, expressed as $\dimInput \times \windowEncWidth$-matrix as 
\begin{equation}
    \windowEncStep_\timeInputWinStep =
    \Big[\inputStepVec_{\windowEncWidth(\timeInputStep'-1)+1},\inputStepVec_{\windowEncWidth(\timeInputStep'-1)+2},\ldots,\inputStepVec_{\windowEncWidth(\timeInputStep'-1)+\windowEncWidth}\Big],
    \label{eq:input-window-matrix}
\end{equation}
and $\repEncStep_\timeInputWinStep$ is extracted representation vector inputted to the seq2seq encoder as shown in Fig.~\ref{fig:wave2wave-architecture}
---the dimension of $\repEncStep$ is the same as the one of the hidden vector
$^{\hiddenEncStep}$.

Similarly, to approximate the complex target waves well, we introduce inverse representation function, 
$\invRepFunc(\cdot)$ which is separately trained from $\invRepFunc(\cdot)$ as
\begin{equation}
   \windowDecStep_\timeOutputWinStep = \invRepFunc(\repDecStep_\timeOutputWinStep)
   \label{eq:inverse-representation}
\end{equation}
where $\repDecStep_\timeOutputWinStep$ is the $\timeOutputWinStep$-th output vector from seq2seq decoder as shown in Fig~\ref{fig:wave2wave-architecture}, and  
$\windowDecStep_\timeOutputWinStep$ is a window matrix which is corresponding to a partial wave of target
waves $\outputSeqVecD = (\outputStepVec_1,\dots, \outputStepVec_\seqOutputLen)$.

The advantage of window-based architecture are three-fold: firstly, the number of steps in both 
encoder and decoder could be largely reduced and make the seq2seq with context vector work stably.
Secondly, the complexity and variation in the shape inside windows are also largely reduced in comparison with the entire waves.
Thus, important information could be extracted from source waves and the output sequence could be accurately approximated
by relatively simple representation $\repFunc(\cdot)$ and inverse-representation $\invRepFunc(\cdot)$ functions respectively.
Thirdly, both representation $\repFunc(\cdot)$ and inverse-representation $\invRepFunc(\cdot)$ functions are
trained end-to-end manner by minimizing the loss $\loss$ where both functions are modeled by fully-connected (FC) networks.

Fig. \ref{fig:wave2wave-architecture} depicts the overall architecture of our wave2wave with an example of toy-data.
The wave2wave consists of encoder and decoder with long-short term memory (LSTM) nodes in their inside, 
representation function $\repFunc(\windowEncStep_\timeInputWinStep)$ and
inverse-representation function $\invRepFunc(\windowDecStep_\timeOutputWinStep)$.
In this figure,  one-dimensional $10000$-time-step continuous time-series
are considered as an input and an output
and the width of window is set to $2000$---
there are $5$ window steps for both encoder and decoder, 
i.e., $\windowEncWidth = \windowDecWidth = 2000$ and $\seqInputWinLen = \seqOutputWinLen = 5$. 
Then, $1 \times 2000$ encoder-window-matrix $\windowEncStep_\timeInputWinStep$ is converted to
$\dimRep$ dimensional encoder-representation vector $\repEncStep_\timeInputWinStep$
by the representation function $\repFunc(\windowEncStep_\timeInputWinStep)$ .
Meanwhile, the output decoder, $\dimRep$ dimensional decoder-representation $\repDecStep_\timeOutputWinStep$,
is converted to $1 \times 2000$ decoder-window-matrix $\windowDecStep_\timeOutputWinStep$
by the inverse representation function $\invRepFunc(\repDecStep_\timeOutputWinStep)$.

\subsection{Wave2wave iterative model}
\label{Wave2waveIterative}

We consider two different ways to implement high-dimensional sensor
data.  Since NMT for machine translation handles embeddings of words,
the straightforward extention to high-dimensional settings uses the
$\dimInput$-dimensional source signal at the same time step as source
embeddings, and the $\dimOutput$-dimensional target signal at the same
time step as target embeddings. We call this an wave2wave
model, i.e. the standard model. Alternatively, we can build $\dimOutput$ independent embeddings
separately for corresponding individual 1-dimensional target signal at
each time step while we use the same $\dimInput$-dimensional source
signal embeddings.  We call this a Wave2WaveIterative model.  We
suppose that the former model would be effective when sensor data are
correlated while the latter model would be effective when sensor data
are independent.  Algorithm 1 shows the latter algorithm.

\begin{algorithm}[H]
 \caption{Wave2waveIterative model}
 \SetAlgoLined
 \KwData{src$_{\dimInput \times S}$, tgt$_{\dimOutput \times T}$, $e_{src}$ $\leftarrow$ $\inputStepVec^{\dimInput}$, $e_{tgt_j}$ $\leftarrow$ ${\outputStepVec_j^{\dimOutput}}$}
     \SetKwProg{Def}{def}{:}{}     
     \Def{trainWave2WaveIterative($e_{src} \times S$, $e_{tgt_j} \times T$)}{              
         \For{j = (1,$\dimOutput$)}{
             f(j) = trainWave2Wave($e_{src} \times S$,$e_{tgt_j}$ $\times T$)\;             
         }
       }
\end{algorithm}
The back-translation is a technique to improve the performance by
bi-directional translation removing the noise under a {\it
  neutral-biased} translation \cite{Hoang18}. We deploy this technique
which we call the wave2wave iterative back-translation model.

\section{Evaluation on real-life data: ground motion translation}
\label{sec:ground-motion-translation}

In this section, we apply our proposed method, wave2wave, to predict a broadband-ground motion from only its long-period motion, caused by the same earthquake. In this section, wave2wave translates one dimensional signal wave into one dimensional signal wave.

Ground motions of earthquakes cause fatal damages on buildings and infrastructures.
Physics-based numerical simulators are used to generate ground motions at a specific place, given the property of earthquake, e.g., location and scale to estimate the damages on buildings and infrastructures~\cite{Iwaki2013}.
However, the motion generated by simulators are limited only long periods, longer than $1$ second due to heavy computational costs, and the lack of detailed knowledge of the subsurface structure.

\begin{table}[t]
\begin{center}
\begin{tabular}[b]{c|cc}
\hline
method & train loss & test loss\\
\hline \hline
simple encoder-decoder $\dimZ = 100$ & 1.13 & 0.53\\
simple encoder-decoder $\dimZ = 500$ & 0.90 & 0.47\\
simple encoder-decoder $\dimZ = 1000$ & 0.41 & 0.63\\
\hline
simple seq2seq $\windowEncWidth=\windowDecStep=500$ & $9.27$ & $2.87$\\
simple seq2seq $\windowEncWidth=\windowDecStep=1000$ & $9.87$ & $2.79$\\
simple seq2seq$\windowEncWidth=\windowDecStep=2000$ & $6.82$ & $2.60$\\
\hline
wave2wave $\windowEncWidth=\windowDecStep=500$ & $0.67$ & $0.44$\\
wave2wave $\windowEncWidth=\windowDecStep=1000$ & $0.17$ & $0.34$\\
wave2wave $\windowEncWidth=\windowDecStep=2000$ & $0.25$ & $0.43$\\
\hline
\end{tabular}
\caption{Mean squared loss of simple encoder-decoder methods, simple seq2seq methods and our wave2wave in earthquake ground motion data}
\label{tab:ground-motion-result}
\end{center}
\end{table}

A large amount of ground motion data have been collected by K(kyosin)-NET over the past 20 years in Japan.
Machine learning approaches would be effective to predict broadband-ground motions including periods less than $1$ second, from simulated long period motions.
From this perspective, we apply our method wave2wave to this problem by setting long-ground motion as an input and broadband-ground motion as an output, with the squared loss function $\loss$.

As for training data, we use $365$ ground motion data collected at the observation station, IBR011, located at Ibaraki prefecture, Japan from Jan. 1, 2000 to Dec. 31, 2017---originally there are $374$ data but $10$ data related Tohoku earthquakes and the source deeper than $300\mathrm{m}$ are removed.
As for testing, we use $9$ ground-motion data of earthquakes occurred at the beginning of 2018. 

\begin{figure}[h]
 \begin{center}
     \subfloat[Example of enveloped ground motion]{ \includegraphics[width=0.8\textwidth]{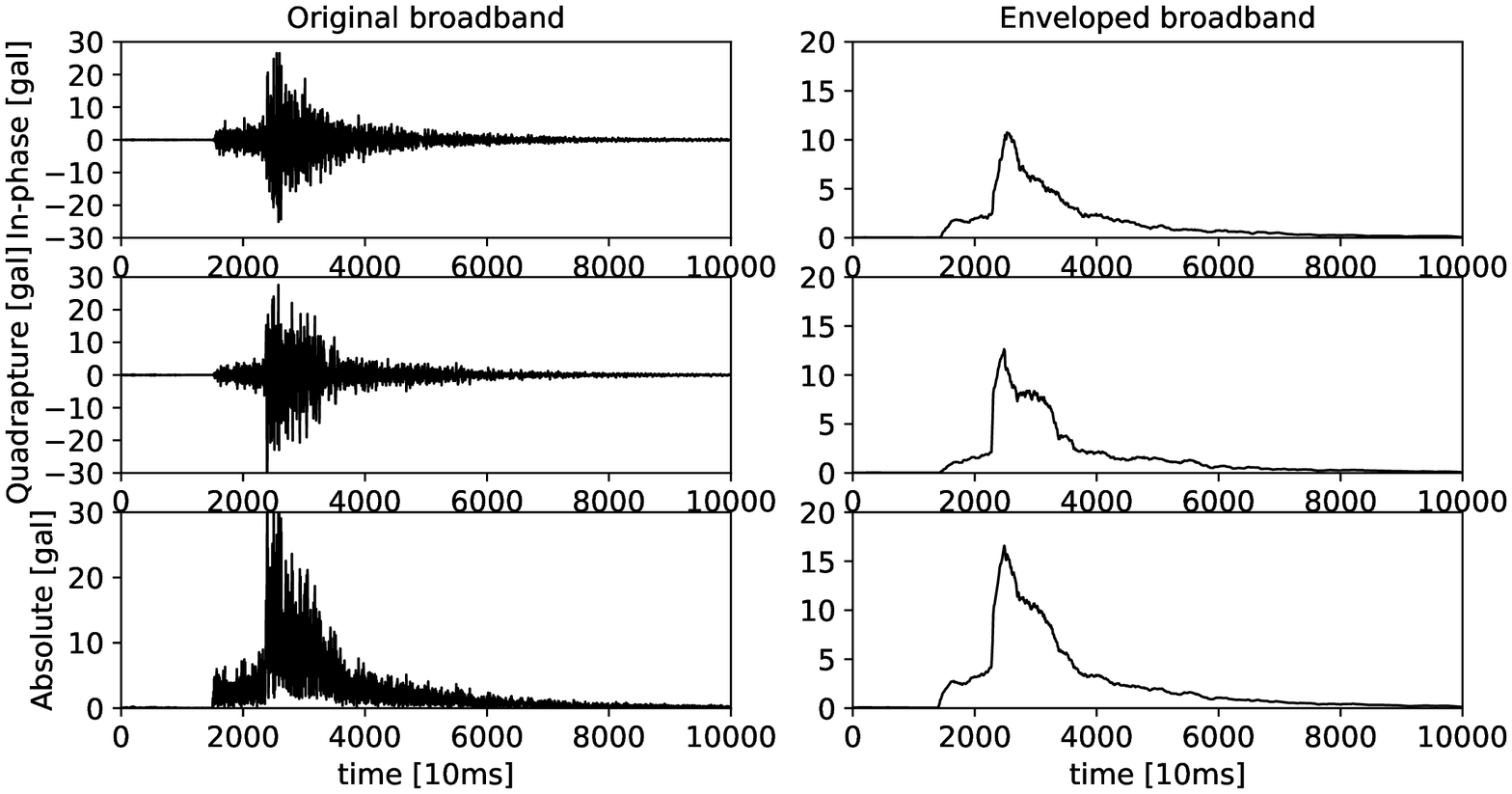}\label{fig:ground-motion-envelop}}
 \end{center}
 \begin{center}
  \subfloat[Predicted broadband ground motion on Jan. 24, 2018 ]{ \includegraphics[width=0.8\textwidth]{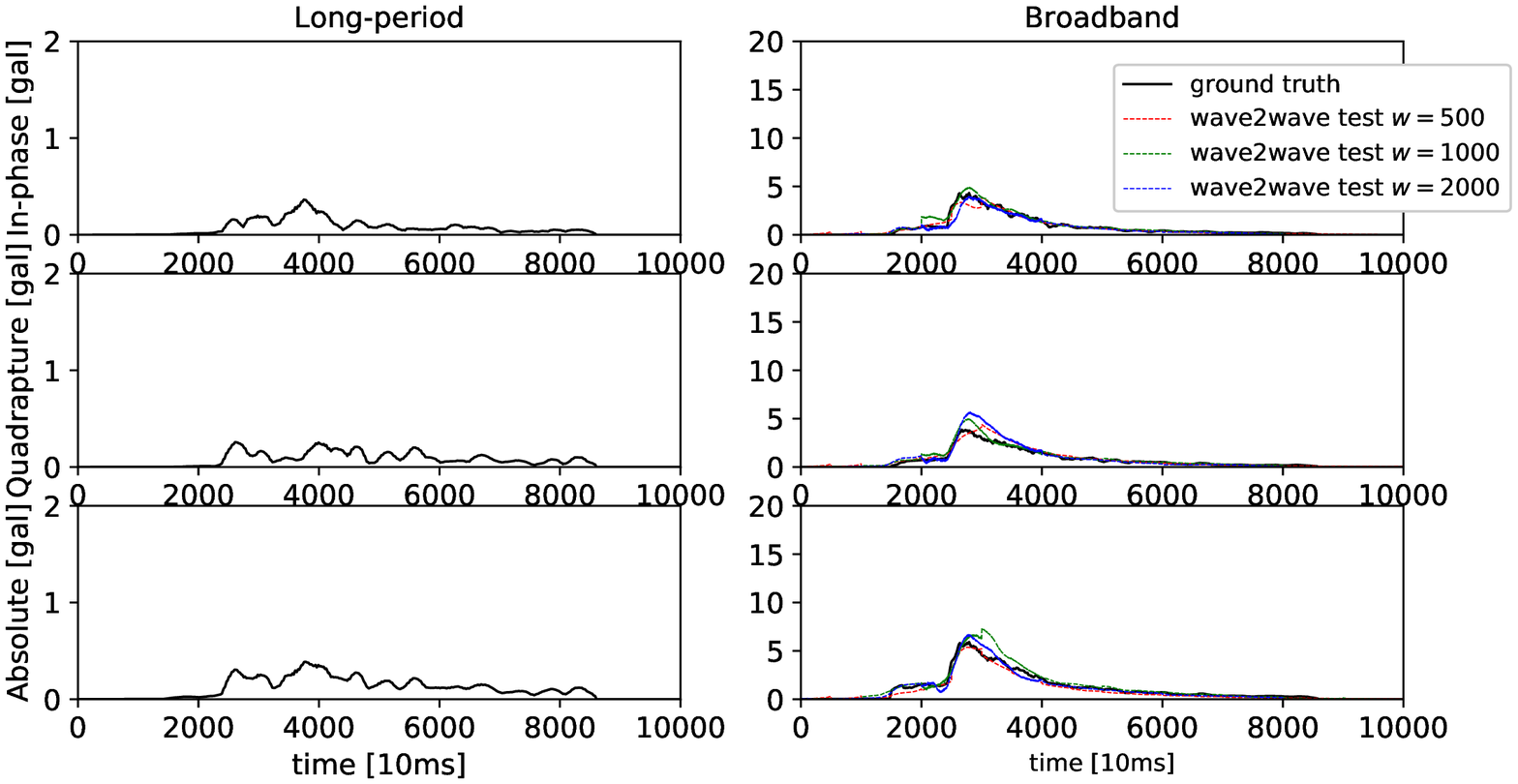}\label{fig:predicted-ground-motion-envelop1}}
 \end{center}
 \begin{center}
  \subfloat[Predicted broadband ground motion on May. 17, 2018 ]{ \includegraphics[width=0.8\textwidth]{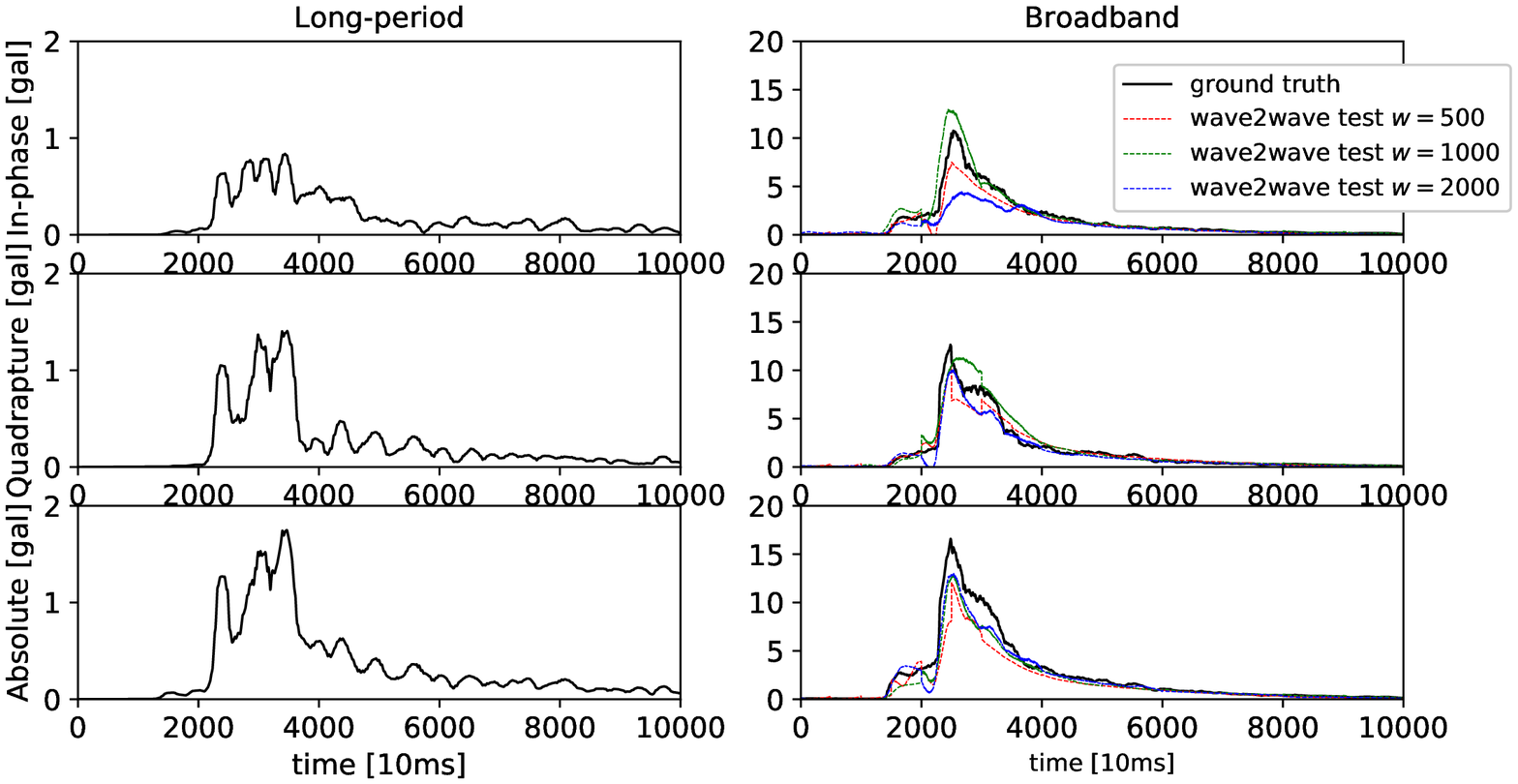}\label{fig:predicted-ground-motion-envelop6}}
 \end{center}
\caption{{\it top}: Example of original and enveloped ground motion data with in-phase, quadrature components and these absolute values. {\it middle and bottom}: predicted broadband ground motion by our methods wave2wave for earthquakes occurred on Jan. 24, 2018 and May. 17, 2018.}
\end{figure}

In addition, both long and broadband ground motion data are cropped to the fixed length, i.e., $\timeInputStep=\timeOutputStep=10000\mathrm{ms}$ and its amplitude is smoothed using RMS (Root Mean Square) envelope with $200\mathrm{ms}$ windows to capture essential property of earthquake motion. 
Moreover, as for data augmentation, in-phase and quadrature components, and those absolute values are extracted from each ground motion. 
That is, there are totally $365 \times 3$ training data.
Fig.~\ref{fig:ground-motion-envelop} shows an example of $3$ components of a ground motion of earthquake occurred on May 17, 2018, Chiba in Japan, and corresponding RMS envelopes.

Table~\ref{tab:ground-motion-result} depict the mean-squared loss of training of three methods, simple encoder-decoder, simple seq2seq, and our proposed method with the same setting as the toy data except $\hiddenEncStep=\hiddenDecStep=50$.
This table shows that our wave2wave methods basically outperform other methods although wave2wave with the small window-width $\windowEncWidth=\windowDecWidth=500$ is lost by simple encoder-decoder with large hidden layer $\dimZ=1000$ in train loss.
This indicates that window-based representation and inverse-representation functions are helpful similarly in toy data.

Fig.~\ref{fig:predicted-ground-motion-envelop1} depicts examples of predicted broadband ground motions of earthquakes occurred on Jan. 24 and May 17, 2018. These show that our method wave2wave predict enveloped broadband ground motion well given long-period ground motion although there is little overfitting due to small training data.

It is expected that predicted broadband-motion combined with simulated long-period motion could be used for more accurate estimation of the damages on buildings and infrastructures.

\vspace{-0.2cm}
\section{Evaluation on real-life data: activity translation}\label{Application}

\begin{figure*}[ht]    
\centering      
     \includegraphics[width=7cm]{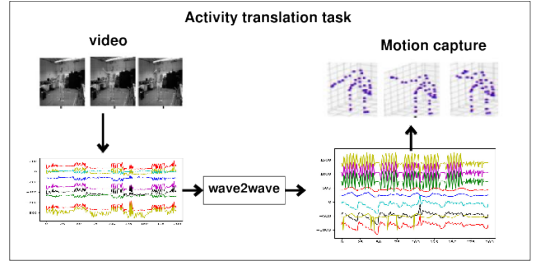}
   \caption{Figure shows activity translation task and activity recognition task which we conduct experiments.}
     \label{overviewActivity}
\end{figure*}

This section deploys wave2wave for activity translation (Refer Fig. \ref{overviewActivity}). Until the previous section, the signals were one dimensions. The signals in this section are high-dimensional in their inputs as well as outputs. 
The dimensions of motion capture, video, and accelerometer are 129, 48, and 18 dimensions, respectively, in
the case of MHAD dataset\footnote{{\tt http://tele-immersion.citris-uc.org/berkeley$\underline{\;\;}$mhad}.}.
Under the mild assumption that the targeted person which are recorded in three different modalities, including motion capture, video, and accelerometer, are synchronized and the noise such as the effect of other surrounding persons is eliminated. Hence, we assume that each signal shows one of the multi-view 
projections of a single person. That is, we can intuitively think that they are equivalent.
Under this condition, we do a translation from motion capture to video (Similarly, accelerometer to motion capture, and video to accelerometer, and these inverse directions).

\vspace{-0.2cm}
\subsubsection{Overall Architecture}

\paragraph{Wave Signal}

Figure \ref{overviewActivity} shows that motion capture and video can be considered as wave signal.
When video is input, $\windowEncStep_\timeInputWinStep$ takes the form of pose vectors which are converted by OpenPose library~\cite{Cao17}. 
Then, this representation is convereted into the window
representation by $\repFunc(\windowEncStep_\timeInputWinStep)$.
When motion capture is input, $\windowEncStep_\timeInputWinStep$ takes the form of motion capture vectors. In this way we used these signals for input as well as output for wave2wave.
The raw output are reconstructed by $\invRepFunc(\windowDecStep_\timeOutputWinStep)$ for the output of representation $\windowDecStep_\timeOutputWinStep$.

\paragraph{Wave Signal Dimensionality Reduction}
As an alternative to use FC layer before the input, we 
use the clustering algorithm, specifically an affinity propagation \cite{Frey07}, in order to reduce the size of representation
as a whole.
While most clustering algorithms need to supply the number of
clusters beforehand, this affinity propagation algorithm solves the
appropriate cluster number as a result. 

\paragraph{Multi-Resolution Spatial Pyramid}
Additinaly structures in wave2wave
is the multi-scalability since the frame rate of multimodal 
data are considerably different.
We adopted the approach of multi-resolution spatial pyramid by a dynamic pose \cite{Neverova14}. We assume that the sequence of frames across
modalities is synchronized and sampled at a given temporal step $v$
and concatenated to form a spatio-temporal 3-d volume.

%\vspace{-1.0cm}

\begin{table}[h]    
\begin{center}
  \begin{tabular}[b]{p{1.9cm}|p{1.5cm}|p{1.8cm}|p{1.5cm}|p{1.8cm}|p{1.5cm}|p{1.8cm}|} \hline    
 ($\windowEncWidth$, $\windowDecStep)$
    & ppl $\dimZ=1$ & ppl $\dimZ=129$ & ppl $\dimZ=1$ & ppl $\dimZ=129$
    & ppl $\dimZ=1$ & ppl $\dimZ=129$
    \\ \hline\hline
    & \multicolumn{2}{c}{seq2seq baseline} & \multicolumn{2}{|c}{seq2seq clustering}& \multicolumn{2}{|c}{} \\\hline\hline              
    & $58000.42$ & $52000.33$ & $5.20$ & $30.22$ & \multicolumn{2}{|c}{} \\ \hline\hline
    & \multicolumn{2}{c}{wave2wave} & \multicolumn{2}{|c}{wave2waveIte} &
\multicolumn{2}{|c}{wave2waveIteBacktrans} \\ \hline\hline
 (1,16) & $2.13$ & $19.74$ & $2.13$ & $4.72$ & 2.13& 4.73\\
 (5,80) & $0.33$ & $10.73$ & $0.33$ & $3.44$  & \underline{$0.32$} & \underline{$3.40$} \\
 (10,160)  & $0.42$ & $11.28$ & $0.42$ & $3.49$ & 0.42& 3.48 \\
 (20,320)  & $0.72$ & $13.67$ & $0.72$ & $3.78$ & 0.72& 3.75 \\
 (30,480)  & $1.21$ & $15.03$ & $1.21$ & $4.11$ & 1.21& 4.11 \\
 (60,960)  & $4.30$ & $35.98$ & $4.30$ & $6.81$ & 4.30& 6.82\\ \hline
  \end{tabular}
  \label{perplexity}
\caption{Figure shows major experimental results for acc2moc.}
\end{center}  
\end{table}

\vspace{-0.3cm}
\subsubsection{Experimental Evaluation}

\paragraph{Experimental Setups}
We used the MHAD dataset from
Berkeley. 
We used video, accelerometer, and mocap
modalities. We used video with Cluster-01/Cam01-02
subsets, and the whole mocap (optical) and accelerometer data with 12
persons/5 trials. 
Video input was
preprocessed by OpenPose which identifies 48 dimensions of vectors.
Optical mocap had the position of the keypoints
whose dimension was 129.  Accelerometer were placed in 6 places
in the body whose dimension was 18. 
We used the parameters in wave2wave with loss function $\calL(\param) = -\frac{1}{\dataNum}\sum_{\dataStep=1}^{\dataNum} $ $\log p_\param (\outputSeqVec^\dataStep|\inputSeq^\dataStep)$
with LSTM modules 500, embedding size 500, dropout 3,
maximum sentence length 400, and batch size 100.  We used
Adam optimizer. We used $v=2, 3, 4$ for multi-resolution spatial
pyramid. We used the same parameter set for wave2wave interactive model.
We use Titan Xp.

\vspace{0.4cm}
\paragraph{Human Understandability}
One characteristic of activity translation can be observed in the
direction of wave2wave translation with accelerometer to video,
e.g. acc2cam. That is, the accelerometer data takes the form that is
not understandable by human beings by its nature but translation to
video makes this visible. By selecting 50 test cases, the human could
understand 48 cases. 96 \% is fairly good.  The second characteristic
of activity translation is opportunistic sensor problem, e.g. when we
cannot use video camera in bathrooms, we use other sensor modality,
e.g. accelerometer, and then translate it to video which can use at
this {\it opportunity}. This corresponds to the case of acceleromter
to video, e.g. acc2cam. We conduct this experiments. Upon watching the
video signals on a screen we could observe the basic human
movements. By selecting 50 test cases, the human could understand 43
cases.

\paragraph{Experimental Results}
Major experimental results are shown in Table 4.  We used
$\windowEncWidth=\{1, 5, 10, 20, 30, 60\}$. For each window size we
measured one target with perplexity (ppl) and the whole target with
perplexity (ppl). We compared several wave2wave results with (1) the
seq2seq model without dimensionality reduction (via clustering), (2)
the seq2seq model with dimensionality reduction. All the experiments
are done with the direction from accelerometer to motion capture
(acc2moc).

Firstly, the original seq2seq model did not work well without
dimensionality reduction of input space. The perplexity was
$58000.42$. This figure suggests that the optimization of deep
learning did not go progress due to the complexity of the training
data or the bad initialization.  However, the results were improved
fairly well if we do dimensionality reduction using clustering. This
figure is close to the results by wave2wave (iterative) with
$\windowEncWidth=60$.

Secondly, $\windowEncWidth=5$ performed better than other window size
for perplexity when $\dimZ=1$. When this became high dimensional, the
wave2wave iterative model performed better than the wave2wave mode:
$3.44$ vs $10.73$ in perplexity. Since motion capture has $\dimZ=129$
dimensions, the representation space becomes $R^{\dimZ}$ when we let
$R$ denote the parameter space of one point in motion
capture. Compared with this the wave2wave iterative model equipped
with the representation space linear with $R$. The wave2wave iterative
model has an advantage in this point. Moreover, the wave2wave
iterative back-translation model made the best score in perplexity
when $\dimZ=1$ as well as $\dimZ=129$.

\section{Conclusion}

We proposed a method to translate between waves {\it wave2wave} with a
sliding window-based mechanism and iterative back-translation model
for high-dimensional data.  Experimental results for two real-life data show that this is positive.
Performance improvements were about 46 \% in test loss for one dimensional case and about
1625 \% in perplexity for high-dimensional case using the iterative back-translation model.

\bibliographystyle{IEEEtranS}
\bibliography{icdm}

\end{document}